\documentclass[runningheads]{llncs}

\usepackage[mobile]{eccv}

\usepackage{eccvabbrv}

\usepackage{graphicx}
\usepackage{booktabs}

\usepackage[accsupp]{axessibility}  

\usepackage[pagebackref,breaklinks,colorlinks,citecolor=eccvblue]{hyperref}

\usepackage{orcidlink}

\newtheorem{myremark}{Remark}

\usepackage{wrapfig}

\begin{document}

\title{FastBO: Fast HPO and NAS with Adaptive Fidelity Identification} 

\titlerunning{FastBO: Fast HPO and NAS with Adaptive Fidelity Identification}

\author{Jiantong Jiang\orcidlink{/0000-0002-1624-3969} \and
Ajmal Mian\orcidlink{0000-0002-5206-3842}}

\authorrunning{J. Jiang and A. Mian}

\institute{The University of Western Australia, Perth WA 6009, Australia \\
\email{\{jiantong.jiang@research.,ajmal.mian@\}uwa.edu.au}}

\maketitle

\begin{abstract}
  Hyperparameter optimization (HPO) and neural architecture search (NAS) are powerful in attaining state-of-the-art machine learning models, with Bayesian optimization (BO) standing out as a mainstream method. Extending BO into the multi-fidelity setting has been an emerging research topic, but faces the challenge of determining an appropriate fidelity for each hyperparameter configuration to fit the surrogate model. To tackle the challenge, we propose a multi-fidelity BO method named FastBO, which adaptively decides the fidelity for each configuration and efficiently offers strong performance. The advantages are achieved based on the novel concepts of \emph{efficient point} and \emph{saturation point} for each configuration. We also show that our adaptive fidelity identification strategy provides a way to extend any single-fidelity method to the multi-fidelity setting, highlighting its generality and applicability.
\end{abstract}

\section{Introduction}
\label{sec:intro}

HPO~\cite{feurer2019hyperparameter} and NAS~\cite{nas} aim to find the hyperparameter configuration or architecture $\boldsymbol{\lambda}^*$ that minimizes $f(\boldsymbol{\lambda})$, the performance obtained by configuration $\boldsymbol{\lambda}$.
BO~\cite{bo-gp, bo-rf, bo-tpe} is an effective model-based method for HPO and NAS. It maintains a \emph{surrogate model} of the performance based on past evaluations of configurations, which guides the choice of promising configurations to evaluate.
Recent studies on BO have explored expert priors~\cite{shahriari2016unbounded, bock, DBLP:conf/icdm/0003RG0VSLSHMG18, pibo}, derivative information~\cite{wu2017bayesian, padidar2021scaling, ament2022scalable}, and enhancing the interpretability~\cite{chen2019looks, yang2022re, yang2023local, yang2024backdoor, yang2024regulating} of HPO and NAS~\cite{bischl2023hyperparameter, moosbauer2021explaining, moosbauer2022improving}.


However, standard BO requires full evaluations of configurations, which incurs significant costs, especially considering the escalating model evaluation overhead. Despite efforts to accelerate model evaluation ~\cite{jiang2022fast, fast-bni, jiang2024fast, jiang2024fastpgm}, smart strategies are required to widely adopt HPO and NAS. 
Thus, multi-fidelity methods have been proposed~\cite{sha, hb, asha, pasha}, where the \emph{fidelities} mean the performance levels obtained under various resource levels. They follow the idea of successive halving (SHA)~\cite{sha}: initially, they evaluate many random configurations using few resources; then, based on the low-fidelity performances, only the well-performing ones successively continue to be evaluated with increasing resources. 

Follow-up studies~\cite{bohb, bohb-like, mob, hypertune, cqr} propose model-based multi-fidelity methods, replacing random sampling with more informed models to improve sample efficiency.
However, they are based on SHA, which assumes that learning curves of different configurations rarely intersect - a condition that often fails in practice~\cite{viering2022shape}, i.e., early performance observations cannot always indicate the final fidelity performance. This raises a key challenge: \emph{What is the appropriate fidelity for each configuration to fit the surrogate model?} In other words, which fidelity can offer observations that reliably indicate the final fidelity performance? Current methods struggle with this.  Hyper-Tune~\cite{hypertune} and BOHB~\cite{bohb} fit separate models for different fidelities, missing inter-fidelity correlations. FTBO~\cite{ftbo} and A-BOHB~\cite{mob} fit joint models but require strong assumptions. Salinas et al.~\cite{cqr} use the last observed fidelity performance, which may get inaccurate surrogate models as it widens the gap between poorly- and well-performing configurations. 

This paper is an extended abstract of our conference paper~\cite{jiang2024efficient}, highlighting key ideas and the main experimental results, while omitting finer details.

\section{Key Idea of FastBO}

We propose a multi-fidelity extension of BO, namely FastBO, which tackles the challenge of deciding the appropriate fidelity for each configuration to fit the surrogate model. Here, we first propose the key concepts of \emph{efficient point} and \emph{saturation point}, which are crucial in the optimization process. Then, we briefly describe the process of FastBO and highlight its generality.

\subsection{Efficient Point and Saturation Point}

We first formally define the efficient point as follows.
\vspace{-0.5mm}
\begin{definition}[Efficient point]
\label{def_point1}
    For a given learning curve $\mathcal{C}_i(r)$ of hyperparameter configuration or architecture $\boldsymbol{\lambda}_i$, where $r$ represents the resource level (also referred to as fidelity), the efficient point $e_i$ of $\boldsymbol{\lambda}_i$ is defined as: $e_i = \min\{r \mid \mathcal{C}_i(r) - \mathcal{C}_i(2r) < \delta_1\}$, where $\delta_1$ is a predefined small threshold.
\end{definition}
\vspace{-0.5mm}
The semantic of Definition~\ref{def_point1} is that starting from the efficient point onwards, when the resources are doubled, the performance improvement falls below a small threshold.
Consequently, this point signifies a fidelity of performance achieved with comparably efficient resource usage. 
Thus, we make the following remark.
\vspace{-0.5mm}
\begin{myremark}
\label{remark_point1}
    The efficient points of the configurations can serve as their appropriate fidelities used for fitting the surrogate model. This is due to their (i) optimal resource-to-performance balance, (ii) ability to capture valuable learning curve trends, and (iii) customization for different hyperparameter configurations. 
\end{myremark}
\vspace{-0.5mm}
We elaborate on the reasons as follows.
Firstly, efficient points balance the trade-off between computational cost and result quality. Beyond the efficient point, allocating additional resources becomes less efficient. 
Secondly, efficient points capture valuable behaviors within the learning curves, enabling more informed decision-making.
Thirdly, the ability to customize the fidelity for each specific configuration is an advantage. This adaptive approach is more reasonable than previous studies that use a fixed fidelity for all the configurations.

Besides efficient points, we identify saturation points for all configurations as well. We provide the definition of the saturation point as follows.
\vspace{-0.5mm}
\begin{definition}[Saturation point]
\label{def_point2}
    For a given learning curve $\mathcal{C}_i(r)$ of configuration $\boldsymbol{\lambda}_i$, where $r$ represents the resource level (also referred to as fidelity), the saturation point $s_i$ of $\boldsymbol{\lambda}_i$ is defined as: $s_i = \min\{r \mid \forall r' > r, |\mathcal{C}_i(r') - \mathcal{C}_i(r)| < \delta_2\}$, where $\delta_2$ is a predefined small threshold.
\end{definition}
\vspace{-0.5mm}
The semantic of Definition~\ref{def_point2} is that beyond the saturation point, the observed performance no longer exhibits notable variations with more resources. Thus, this point characterizes the fidelity at which the performance of a configuration stabilizes. 
Building on the above definition, we make the following remark.
\vspace{-0.5mm}
\begin{myremark}
\label{remark_point2}
    The saturation points of the configurations can serve as their approximate final fidelities, as they provide performance results that meet predefined quality thresholds while reducing resource wastage. 
\end{myremark}
\vspace{-0.5mm}
%

\begin{figure*} [tb]
	\centering
        \vspace{-2mm}
	\includegraphics[width=1.0\textwidth]{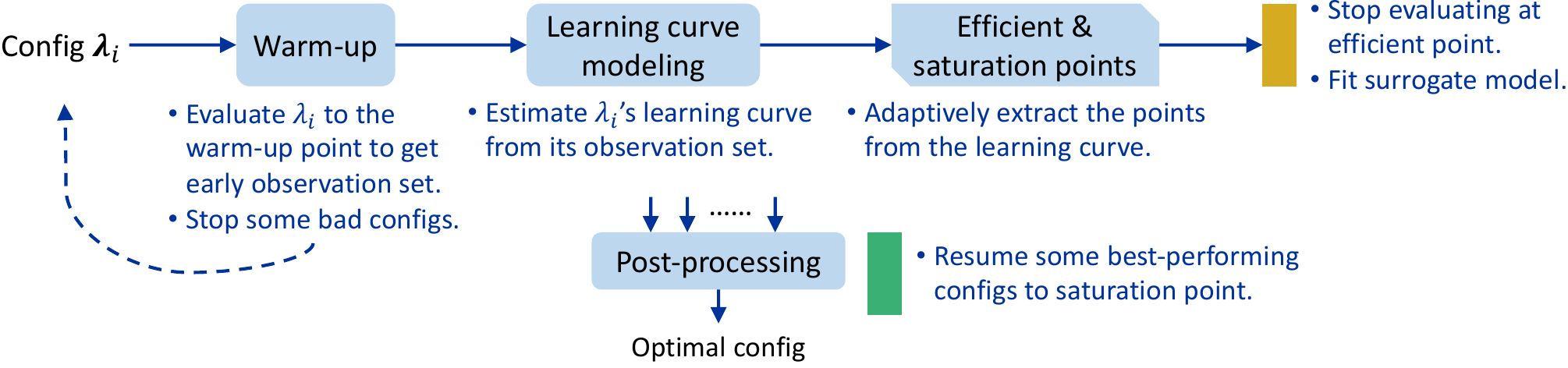}
	\caption{Main process of FastBO. FastBO involves estimating efficient and saturation points, modeling learning curves, and auxiliary stages of warm-up and post-processing.}
	\label{fig_process}
        \vspace{-4mm}
\end{figure*}
\subsection{FastBO Process and Generalization}

With the two crucial points, we show the main process of FastBO in Fig.~\ref{fig_process}.
Each configuration $\boldsymbol{\lambda}_i$ first enters a warm-up stage to get its early observation set. Some configurations are terminated here if they are detected consecutive performance deterioration.
Then, FastBO estimates the learning curve of $\boldsymbol{\lambda}_i$ from its observation set.
Thus, the efficient point and saturation points are adaptively extracted. 
After that, $\boldsymbol{\lambda}_i$ continues to be evaluated to its efficient point; the result is used to update the surrogate model.
Finally, the post-processing stage let a small set of promising configurations resume evaluating to their saturation points, and the optimal configurations can be obtained.

\noindent\textbf{Generalizing FastBO to single-fidelity methods.}
The inefficiency of single-fidelity methods like BO stems from their reliance on expensive final fidelity evaluations. Notably, low-fidelity evaluations provide informative insights but are computationally cheaper. Therefore, we can extend single-fidelity methods to the multi-fidelity setting by acquiring the low-fidelity performance for each configuration to fit the surrogate model. 
To do this, we need to determine the fidelity used to fit the surrogate model. FastBO adaptively determines this fidelity for each configuration by identifying its efficient point. While this adaptive identification strategy is described in the context of model-based methods, it can be generalized to various single-fidelity methods. For example, when evaluating configurations within the population for an evolutionary algorithm-based HPO method, we can similarly evaluate the efficient point performances instead of the final performances and integrate them in the subsequent processes, such as selection and variation. To conclude, the proposed strategy provides a simple but effective way to bridge the gap between single- and multi-fidelity methods.

\section{Experimental Evaluation}
\begin{figure*}[t]
\centerline{\includegraphics[width=1.0\textwidth]{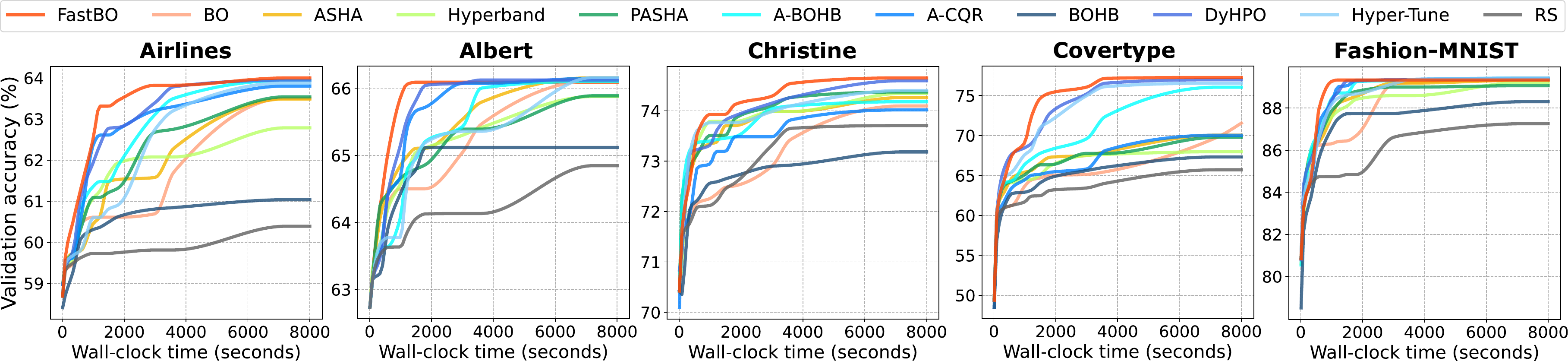}}
	\centering
	\caption{Anytime performance on the LCBench benchmark.}
\label{fig_exp_lcbench}
\vspace{-2mm}
\end{figure*}

\begin{figure*}[t]
\centerline{\includegraphics[width=1.0\textwidth]{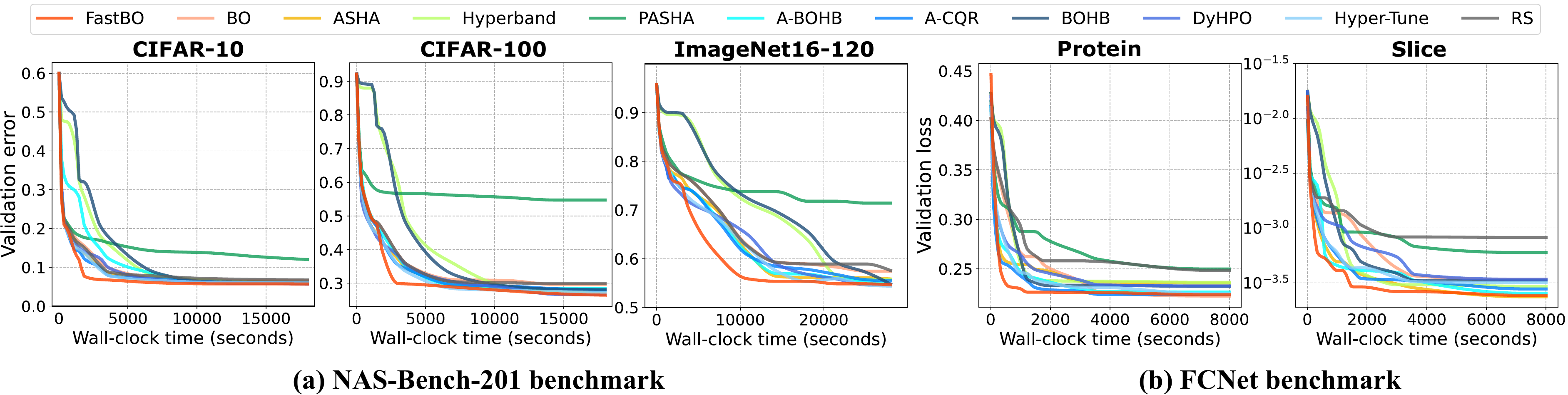}}
	\centering
	\caption{Anytime performance on \textbf{(a)} NAS-Bench-201 and \textbf{(b)} FCNet.}
\label{fig_exp_nas201_fcnet}
\vspace{-4mm}
\end{figure*}
We compare the performance of FastBO with random search (RS), standard BO~\cite{bo-gp}, ASHA~\cite{asha}, Hyperband~\cite{hb}, PASHA~\cite{pasha}, A-BOHB~\cite{mob}, A-CQR~\cite{cqr}, BOHB~\cite{bohb}, DyHPO~\cite{dyhpo}, and Hyper-Tune~\cite{hypertune}. 
The results on the LCBench~\cite{lcbench}, NAS-Bench-201~\cite{nasbench201}, and FCNet~\cite{fcnet} benchmarks are shown in Figs.~\ref{fig_exp_lcbench} and~\ref{fig_exp_nas201_fcnet}. 
Overall, FastBO can handle various performance metrics and shows strong anytime performance. We can observe that FastBO gains an advantage earlier than other methods, rapidly converging to the global optimum after the initial phase.

\section{Conclusion and Discussion}

We propose FastBO, a model-based multi-fidelity HPO method, which excels in adaptively identifying the fidelity for each configuration to fit the surrogate model and efficiently providing high-quality performance. The proposed adaptive fidelity identification strategy also provides a simple way to extend any single-fidelity method to the multi-fidelity setting. 
While this paper provides a strong foundation on HPO and NAS, we see challenges that demand future improvements. Future work could refine and expand Fast-BO to larger search spaces and distributed computing systems to improve its applicability and scalablity.





%
%
\bibliographystyle{splncs04}
\bibliography{main}

\begin{thebibliography}{10}
\providecommand{\url}[1]{\texttt{#1}}
\providecommand{\urlprefix}{URL }
\providecommand{\doi}[1]{https://doi.org/#1}

\bibitem{ament2022scalable}
Ament, S.E., Gomes, C.P.: Scalable first-order {B}ayesian {O}ptimization via structured automatic differentiation. In: International Conference on Machine Learning. pp. 500--516. PMLR (2022)

\bibitem{bo-tpe}
Bergstra, J., Bardenet, R., Bengio, Y., K{\'e}gl, B.: Algorithms for hyper-parameter optimization. Advances in Neural Information Processing Systems  \textbf{24} (2011)

\bibitem{bischl2023hyperparameter}
Bischl, B., Binder, M., Lang, M., Pielok, T., Richter, J., Coors, S., Thomas, J., Ullmann, T., Becker, M., Boulesteix, A.L., et~al.: Hyperparameter optimization: Foundations, algorithms, best practices, and open challenges. Wiley Interdisciplinary Reviews: Data Mining and Knowledge Discovery  \textbf{13}(2),  e1484 (2023)

\bibitem{pasha}
Bohdal, O., Balles, L., Wistuba, M., Ermis, B., Archambeau, C., Zappella, G.: {PASHA:} efficient {HPO} and {NAS} with progressive resource allocation. In: International Conference on Learning Representations. OpenReview.net (2023)

\bibitem{chen2019looks}
Chen, C., Li, O., Tao, D., Barnett, A., Rudin, C., Su, J.K.: This looks like that: deep learning for interpretable image recognition. Advances in Neural Information Processing Systems  \textbf{32} (2019)

\bibitem{nasbench201}
Dong, X., Yang, Y.: {NAS-Bench-201}: Extending the scope of reproducible neural architecture search. In: International Conference on Learning Representations (2020)

\bibitem{nas}
Elsken, T., Metzen, J.H., Hutter, F.: Neural architecture search: A survey. The Journal of Machine Learning Research  \textbf{20}(1),  1997--2017 (2019)

\bibitem{bohb}
Falkner, S., Klein, A., Hutter, F.: {BOHB}: Robust and efficient hyperparameter optimization at scale. In: International Conference on Machine Learning. pp. 1437--1446. PMLR (2018)

\bibitem{feurer2019hyperparameter}
Feurer, M., Hutter, F.: Hyperparameter optimization. Automated Machine Learning: Methods, Systems, Challenges pp. 3--33 (2019)

\bibitem{bo-rf}
Hutter, F., Hoos, H.H., Leyton-Brown, K.: Sequential model-based optimization for general algorithm configuration. In: Learning and Intelligent Optimization. pp. 507--523. Springer (2011)

\bibitem{pibo}
Hvarfner, C., Stoll, D., Souza, A.L.F., Lindauer, M., Hutter, F., Nardi, L.: {\textdollar}{\textbackslash}pi{\textdollar}{BO}: Augmenting acquisition functions with user beliefs for bayesian optimization. In: International Conference on Learning Representations. OpenReview.net (2022)

\bibitem{sha}
Jamieson, K., Talwalkar, A.: Non-stochastic best arm identification and hyperparameter optimization. In: Artificial Intelligence and Statistics. pp. 240--248. PMLR (2016)

\bibitem{fast-bni}
Jiang, J., Wen, Z., Mansoor, A., Mian, A.: Fast parallel exact inference on {B}ayesian networks. In: ACM SIGPLAN Annual Symposium on Principles and Practice of Parallel Programming. pp. 425--426 (2023)

\bibitem{jiang2024efficient}
Jiang, J., Wen, Z., Mansoor, A., Mian, A.: Efficient hyperparameter optimization with adaptive fidelity identification. In: Proceedings of the IEEE/CVF Conference on Computer Vision and Pattern Recognition. pp. 26181--26190 (2024)

\bibitem{jiang2024fast}
Jiang, J., Wen, Z., Mansoor, A., Mian, A.: Fast inference for probabilistic graphical models. In: 2024 USENIX Annual Technical Conference (USENIX ATC 24) (2024)

\bibitem{jiang2022fast}
Jiang, J., Wen, Z., Mian, A.: Fast parallel bayesian network structure learning. In: IEEE International Parallel and Distributed Processing Symposium. pp. 617--627. IEEE (2022)

\bibitem{jiang2024fastpgm}
Jiang, J., Wen, Z., Yang, P., Mansoor, A., Mian, A.: Fast-pgm: Fast probabilistic graphical model learning and inference. arXiv preprint arXiv:2405.15605  (2024)

\bibitem{fcnet}
Klein, A., Hutter, F.: Tabular benchmarks for joint architecture and hyperparameter optimization. arXiv preprint arXiv:1905.04970  (2019)

\bibitem{mob}
Klein, A., Tiao, L.C., Lienart, T., Archambeau, C., Seeger, M.: Model-based asynchronous hyperparameter and neural architecture search. arXiv preprint arXiv:2003.10865  (2020)

\bibitem{DBLP:conf/icdm/0003RG0VSLSHMG18}
Li, C., Rana, S., Gupta, S., Nguyen, V., Venkatesh, S., Sutti, A., de~Celis~Leal, D.R., Slezak, T., Height, M., Mohammed, M., Gibson, I.: Accelerating experimental design by incorporating experimenter hunches. In: International Conference on Data Mining. pp. 257--266. {IEEE} Computer Society (2018)

\bibitem{asha}
Li, L., Jamieson, K., Rostamizadeh, A., Gonina, E., Ben-Tzur, J., Hardt, M., Recht, B., Talwalkar, A.: A system for massively parallel hyperparameter tuning. Proceedings of Machine Learning and Systems  \textbf{2},  230--246 (2020)

\bibitem{hb}
Li, L., Jamieson, K., DeSalvo, G., Rostamizadeh, A., Talwalkar, A.: Hyperband: A novel bandit-based approach to hyperparameter optimization. Journal of Machine Learning Research  \textbf{18}(1),  6765--6816 (2017)

\bibitem{hypertune}
Li, Y., Shen, Y., Jiang, H., Zhang, W., Li, J., Liu, J., Zhang, C., Cui, B.: Hyper-tune: Towards efficient hyper-parameter tuning at scale. Proceedings of the VLDB Endowment  \textbf{15}(6),  1256--1265 (2022)

\bibitem{moosbauer2022improving}
Moosbauer, J., Casalicchio, G., Lindauer, M., Bischl, B.: Improving accuracy of interpretability measures in hyperparameter optimization via {B}ayesian algorithm execution. arXiv preprint arXiv:2206.05447  (2022)

\bibitem{moosbauer2021explaining}
Moosbauer, J., Herbinger, J., Casalicchio, G., Lindauer, M., Bischl, B.: Explaining hyperparameter optimization via partial dependence plots. Advances in Neural Information Processing Systems  \textbf{34},  2280--2291 (2021)

\bibitem{bock}
Oh, C., Gavves, E., Welling, M.: {BOCK}: Bayesian optimization with cylindrical kernels. In: International Conference on Machine Learning. pp. 3868--3877. PMLR (2018)

\bibitem{padidar2021scaling}
Padidar, M., Zhu, X., Huang, L., Gardner, J., Bindel, D.: Scaling gaussian processes with derivative information using variational inference. Advances in Neural Information Processing Systems  \textbf{34},  6442--6453 (2021)

\bibitem{cqr}
Salinas, D., Golebiowski, J., Klein, A., Seeger, M.W., Archambeau, C.: Optimizing hyperparameters with conformal quantile regression. In: International Conference on Machine Learning. vol.~202, pp. 29876--29893. {PMLR} (2023)

\bibitem{shahriari2016unbounded}
Shahriari, B., Bouchard-C{\^o}t{\'e}, A., Freitas, N.: Unbounded {B}ayesian {O}ptimization via regularization. In: Artificial intelligence and statistics. pp. 1168--1176. PMLR (2016)

\bibitem{bo-gp}
Snoek, J., Larochelle, H., Adams, R.P.: Practical {B}ayesian optimization of machine learning algorithms. Advances in Neural Information Processing Systems  \textbf{25} (2012)

\bibitem{ftbo}
Swersky, K., Snoek, J., Adams, R.P.: Freeze-thaw {B}ayesian optimization. arXiv preprint arXiv:1406.3896  (2014)

\bibitem{viering2022shape}
Viering, T., Loog, M.: The shape of learning curves: a review. IEEE Transactions on Pattern Analysis and Machine Intelligence  (2022)

\bibitem{bohb-like}
Wang, J., Xu, J., Wang, X.: Combination of hyperband and bayesian optimization for hyperparameter optimization in deep learning. arXiv preprint arXiv:1801.01596  (2018)

\bibitem{dyhpo}
Wistuba, M., Kadra, A., Grabocka, J.: Supervising the multi-fidelity race of hyperparameter configurations. Advances in Neural Information Processing Systems  \textbf{35},  13470--13484 (2022)

\bibitem{wu2017bayesian}
Wu, J., Poloczek, M., Wilson, A.G., Frazier, P.: Bayesian optimization with gradients. Advances in neural information processing systems  \textbf{30} (2017)

\bibitem{yang2024backdoor}
Yang, P., Akhtar, N., Jiang, J., Mian, A.: Backdoor-based explainable ai benchmark for high fidelity evaluation of attribution methods. arXiv preprint arXiv:2405.02344  (2024)

\bibitem{yang2024regulating}
Yang, P., Akhtar, N., Shah, M., Mian, A.: Regulating model reliance on non-robust features by smoothing input marginal density. arXiv preprint arXiv:2407.04370  (2024)

\bibitem{yang2023local}
Yang, P., Akhtar, N., Wen, Z., Mian, A.: Local path integration for attribution. In: Proceedings of the AAAI Conference on Artificial Intelligence. vol.~37, pp. 3173--3180 (2023)

\bibitem{yang2022re}
Yang, P., Akhtar, N., Wen, Z., Shah, M., Mian, A.: Re-calibrating feature attributions for model interpretation. In: International Conference on Learning Representations (2022)

\bibitem{lcbench}
Zimmer, L., Lindauer, M.T., Hutter, F.: {A}uto-{P}ytorch: Multi-fidelity metalearning for efficient and robust {AutoDL}. IEEE Transactions on Pattern Analysis and Machine Intelligence  \textbf{43},  3079--3090 (2021)

\end{thebibliography}
\end{document}